\documentclass[10pt,twocolumn,letterpaper]{article}

\usepackage{cvpr}              
\usepackage{CJKutf8}
\usepackage{graphicx}
\usepackage{bm}
\usepackage{tabularx}
\usepackage{threeparttable}
\usepackage{multirow}
\usepackage{xcolor}
\usepackage{colortbl}
\usepackage{comment}
\usepackage{pifont}
\usepackage{makecell}
\definecolor{cvprblue}{rgb}{0.21,0.49,0.74}
\usepackage[pagebackref,breaklinks,colorlinks,allcolors=cvprblue]{hyperref}

\def\paperID{} 
\def\confName{CVPR}
\def\confYear{2026}

\title{Breaking Degradation Coupling: A Structural Entropy–Guided Decoupled Framework and Benchmark for Infrared Enhancement}

\author{
	Pu Li$^{1}$ \quad
    Huafeng Li$^{1}$ \quad
	Yafei Zhang$^{1}$ \quad
	Yu Liu$^{2}$\quad
	Wen Wang$^{3}$\thanks{Corresponding author: Wen Wang (sophia\_84@126.com)}\\
	$^{1}$Faculty of Information Engineering and Automation, Kunming University of Science and Technology\\
	$^{2}$Department of Biomedical Engineering, Hefei University of Technology\\
	$^{3}$School of Mathematics and Statistics, Yunnan University\\
}

\begin{document}
\maketitle
\begin{abstract}
Thermal infrared image enhancement aims to restore high-quality images from complex compound degradations. Existing all-in-one approaches typically employ a single shared backbone to handle diverse degradations, which causes gradient interference and parameter competition. To address this, we propose a Structural Entropy–Guided Decoupled (SEGD) Framework. Unlike unified modeling paradigms, SEGD decomposes compound degradations into independent sub-processes and models them in a divide-and-conquer manner through Degradation-Specific Residual Modules (DRMs). Each DRM focuses on residual estimation for a specific degradation, enabling task decoupling while remaining jointly trainable, which mitigates parameter contention. A Degradation-Aware Evidential Network further estimates degradation type and intensity, providing priors that adaptively regulate DRM restoration strength. To handle compound cases, DRMs are composed in varying orders to form multiple restoration paths, from which the most informative features are aggregated under a structural-entropy criterion, yielding decoder-ready representations with structural fidelity and degradation awareness. Integrating divide-and-conquer restoration, evidential perception, and entropy-guided adaptation, SEGD achieves fine-grained and interpretable enhancement. We also construct a nighttime TIR benchmark for evaluation under real low-light conditions. Experimental results demonstrate that SEGD surpasses state-of-the-art methods while achieving higher efficiency with fewer parameters. The source code is available at \url{https://github.com/Kust-lp/SEGD}.
\end{abstract}    
\section{Introduction}
\label{sec:intro}

Thermal infrared (TIR) image enhancement aims to restore high-quality images from observations degraded by noise, blur, and low contrast. In practice, TIR images often experience compound degradations arising from the coupling of multiple factors, including limited scene temperature contrast, atmospheric transmission effects (absorption, scattering, and turbulence), optical defocus, aperture limitation, and sensor-induced noise~\cite{liu2025deal}. These degradations differ in their physical mechanisms and feature distributions, and their interactions are inherently nonlinear and uncertain~\cite{he2018single}, making model learning more challenging. 

Most existing all-in-one enhancement approaches~\cite{pang2023infrared, liu2025deal, liu2025enhancing} employ a single shared backbone to jointly model multiple degradations. However, such unified modeling neglects the intrinsic differences among degradation types and often suffers from gradient conflicts and parameter competition—different degradations drive inconsistent optimization directions, leading to poor generalization in compound degradation scenarios. Furthermore, there often exists a potential order dependency among degradations, where an inappropriate restoration order can distort subsequent degradation distributions, causing error accumulation and artifact amplification. Hence, achieving decoupled modeling across degradation types and adaptive restoration path selection remains a key challenge in improving enhancement quality.

\begin{figure}[t]
    \centering
    \includegraphics[width=0.95\linewidth, trim={0.8cm 24.5cm 10cm 1cm}, clip]{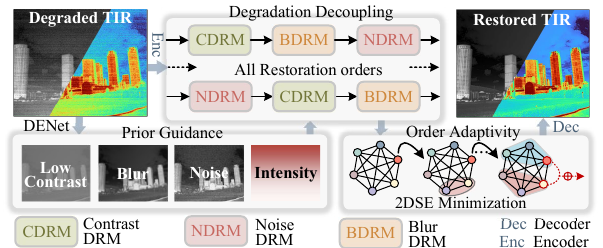}
    \caption{Conceptual illustration of the proposed SEGD framework, which embodies the design philosophy of ``degradation decoupling–prior guidance–order adaptivity''.}
    \label{fig:task}
     \vspace{-6mm}
\end{figure}

To this end, we propose the \textbf{S}tructural \textbf{E}ntropy Guided-Decoupled (\textbf{SEGD}) Framework. 
As shown in Figure~\ref{fig:task}, SEGD follows the design philosophy of ``degradation decoupling--prior guidance--order adaptivity'' and establishes interpretable restoration relationships among multiple degradation types within an information-theoretic modeling paradigm. In our framework, the Degradation-Aware Evidence Network (\textbf{DENet}), grounded in Evidential Deep Learning (EDL)~\cite{sensoy2018evidential}, jointly models degradation type and intensity with uncertainty-aware evidential estimation, constructing explicit and reliable degradation priors that endow the model with degradation awareness. The Degradation-Specific Residual Modules (\textbf{DRMs}) adopt degradation-specific residual learning within independent parameter spaces to realize divide-and-conquer restoration, effectively alleviating gradient interference and parameter competition across degradation types. The Structural Entropy--Guided Restoration Order Selection (\textbf{SE-ROS}) characterizes order dependencies from an information-theoretic perspective, adaptively selecting the most informative restoration paths by minimizing the two-dimensional structural entropy (2D-SE)~\cite{li2016structural,xian2025community}, thereby achieving simultaneous order optimization and feature aggregation. By integrating degradation awareness and order modeling within an optimizable framework, 
SEGD enables fine-grained handling of complex degradation coupling, achieving high-fidelity, low-artifact TIR enhancement. Additionally, we construct a nighttime TIR benchmark, \textbf{Night-TIR}, to evaluate degradation restoration under low-light conditions. 
The main contributions of this work are summarized as follows:
\begin{itemize}
    \item We propose a degradation-aware TIR enhancement framework, which follows the principle of ``degradation decoupling--prior guidance--order adaptivity.'' It establishes a unified restoration pipeline from degradation identification to restoration decision-making, effectively mitigating interference and order uncertainty in compound degradations.

    \item We design two key components: the DENet and DRMs. DENet jointly models degradation type and intensity confidence to generate reliable degradation priors, while DRMs perform divide-and-conquer restoration within independent parameter spaces, reducing gradient conflicts and parameter competition.

    \item We present the SE-ROS, which models order dependencies among degradations and adaptively selects informative restoration paths via 2D-SE minimization, achieving interpretable and unsupervised order optimization.

    \item We construct a nighttime TIR test benchmark (Night-TIR) with physically consistent compound degradations including noise, blur, and low contrast, providing a realistic and challenging testbed for TIR enhancement.
\end{itemize}

\section{Related Work}
\label{sec:relatedwork}

\subsection{Multiple Degraded Visible Image Enhancement}
Multiple degraded visible image enhancement aims to restore clear images from complex compound degradations using a single unified model. Recent studies have proposed various effective approaches~\cite{ma2023prores, DACLIP2024,conde2024instructir,ai2024multimodal,zamfir2025complexity,guo2024onerestore,zheng2024selective,li2023prompt,mao2024allrestorer,hu2025universal,liu2025vl,ren2025manifold}, which can be broadly grouped into three categories: prompt-based, vision large model (VLM)-based, and diffusion-based methods. Prompt-based methods~\cite{ma2023prores,conde2024instructir,ai2024multimodal,li2023prompt,guo2024onerestore} leverage degradation-aware prompts to guide restoration. ProRes~\cite{ma2023prores} encodes multiple degradations into unified visual prompts for controllable enhancement. PIP~\cite{li2023prompt} jointly learns semantic- and texture-level prompts and modulates degradation information via selective interaction. OneRestore~\cite{guo2024onerestore} fuses scene text and image features through cross-attention and introduces a compound degradation loss. InstructIR~\cite{conde2024instructir} employs natural language instructions for all-in-one restoration, removing manually defined prompts. AllRestorer~\cite{mao2024allrestorer} integrates image–text embeddings and designs an All-in-One Transformer Block (AiOTB) for intensity-adaptive enhancement.

VLM-based methods~\cite{DACLIP2024,liu2025vl} exploit the representation power of vision-language models. 
DACLIP~\cite{DACLIP2024} adds a trainable image controller to frozen CLIP~\cite{radford2021learning} to generate high-quality embeddings and reduce representation mismatch. VL-UR~\cite{liu2025vl} extracts semantically aligned embeddings from CLIP as conditional inputs, improving generalization under complex degradations. Diffusion-based methods~\cite{zheng2024selective,ai2024multimodal} adopt clean-image diffusion priors for unified multi-degradation restoration. MPerceiver~\cite{ai2024multimodal} maps CLIP embeddings to textual conditions and extracts multi-scale visual prompts from the latent space of Stable Diffusion~\cite{rombach2022high}. DiffUIR~\cite{zheng2024selective} aligns shared degradation statistics via a selective hourglass mapping strategy, improving restoration fidelity.
Beyond these paradigms, MoCE~\cite{zamfir2025complexity} employs dynamically routed expert modules with varying complexity to adapt to diverse degradations. DCPT~\cite{hu2025universal} pretrains an encoder on degradation classification to inject degradation-understanding priors. MIRAGE~\cite{ren2025manifold} decouples image representations into semantic branches and aligns them on a Symmetric Positive Definite manifold for more robust geometry and stability. However, due to inherent differences in imaging mechanisms and degradation patterns between TIR and visible images~\cite{bao2024thermal,harris1999nonuniformity}, these methods cannot be directly transferred to TIR enhancement.

\subsection{TIR Image Enhancement}
TIR imaging captures object-emitted radiation, enabling reliable perception under nighttime, low-light, and hazy conditions. However, due to atmospheric turbulence, optical defocus, and sensor noise, TIR images often suffer from low contrast, blur, and noise~\cite{liu2025deal}. Many works have addressed these issues through task-specific methods, achieving notable results in single-degradation tasks such as contrast enhancement~\cite{wang2006real,wang2025thermal,qiu2024infrared, wang2021target}, deblurring~\cite{yi2023hctirdeblur,jiang2021thermal,zhang2022combined,varghese2022fast,zhou2023thermal}, and denoising~\cite{liu2019simultaneous,cai2024exploring,hu2023infrared,jiang2025infrared,munch2009stripe,chang2016remote}. Yet, these approaches are designed for isolated degradations and fail to model their joint interactions, leading to performance drops under compound degradations.

Compared with visible image enhancement, research on TIR multi-degradation enhancement remains limited. TSIRIE~\cite{pang2023infrared} uses a dual-stream CNN that combines an infrared detail subnet and a global content subnet to enhance targets and suppress background noise. DEAL~\cite{liu2025deal} formulates enhancement as a two-stage GAN-based process of degradation generation and restoration, reducing reliance on large datasets. PPFN~\cite{liu2025enhancing} introduces degradation-context guidance and selective progressive training to adaptively handle compound degradations. However, these methods typically share one backbone for all degradations, ignoring intrinsic differences, which induces gradient conflicts and parameter competition, limiting performance.

In contrast, the proposed SEGD adopts a divide-and-conquer strategy that decomposes compound degradations into independent sub-processes, each dedicated to a specific type. 
SEGD perceives both degradation type and intensity as priors to guide restoration and employs a structural-entropy minimization criterion to adaptively select the optimal restoration paths in an interpretable manner, substantially improving the visual quality and structural consistency of TIR images under compound degradations.

\section{Methodology}
\label{sec:method}
\begin{figure*}[t]
    \centering
    \includegraphics[width=0.98\linewidth, trim={1cm 18.5cm 1cm 1.3cm}, clip]{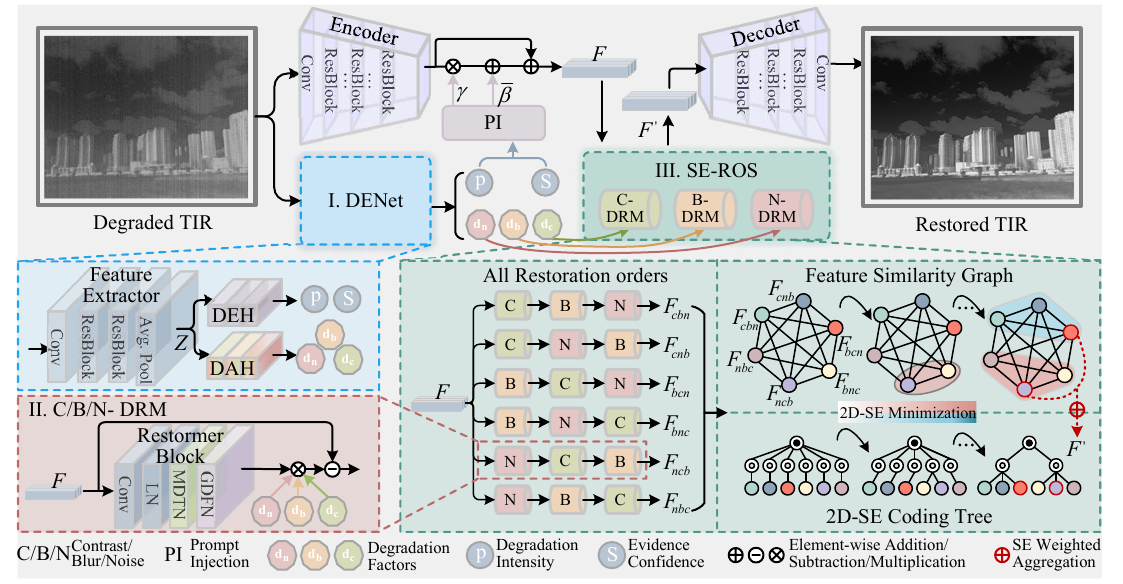} \vspace{-2mm}
    \caption{Architecture of the proposed SEGD framework. SEGD integrates degradation perception, residual restoration, and order-adaptive aggregation in a unified design. DENet estimates degradation type and intensity priors, DRMs conduct degradation-specific residual restoration, and SE-ROS adaptively selects the optimal restoration orders via 2D SE minimization.}\vspace{-3mm}
    \label{fig:model}
\end{figure*}

\subsection{Overview}
As illustrated in Figure~\ref{fig:model}, the proposed SEGD framework comprises three core components:  (1) The DENet explicitly estimates degradation type and intensity and uses them as degradation priors to dynamically modulate the restoration strength of the DRMs.  (2) Guided by these priors, the DRMs independently reconstruct features corresponding to noise, blur, and low-contrast degradations, generating degradation-free representations and effectively mitigating gradient interference and parameter competition across different degradation processes.  (3) The SE-ROS constructs candidate feature similarity graph along multiple restoration paths and adaptively selects the most informative features for aggregation by minimizing 2D-SE, producing high-quality, decoder-ready representation.  Through the synergistic mechanism of divide-and-conquer learning, prior-guided restoration, and order-adaptive aggregation, SEGD enables fine-grained and interpretable modeling of complex compound degradations, achieving notable improvements in both visual quality and structural fidelity.

\subsection{Degradation-Aware Evidence Network}
\label{sec:DAEC}
Compound degradations exhibit high diversity in both type and intensity. Neglecting the specific priors associated with individual degradation types can easily lead to over-restoration or under-restoration. To address this issue, we propose the DENet, which explicitly perceives degradation type and intensity and uses them as degradation priors to dynamically guide the restoration behavior of each DRM. 
This design prevents unnecessary restoration for absent degradations and adaptively adjusts the restoration strength according to degradation severity. As illustrated in Figure~\ref{fig:model}(I), DENet consists of a feature extractor, a Degradation-Aware Head (DAH), and a Degradation-Evidence Head (DEH). Given a degraded image ${\bf I}^d$, the feature extractor first produces a latent representation:
\begin{equation}\small
{\bf Z} = \mathrm{FE}({\bf I}^d),
\end{equation}
where $\mathrm{FE}(\cdot)$ denotes the feature extraction network composed of multiple convolutional layers and residual blocks (ResBlocks). The DAH employs a multilayer perceptron to predict the degradation types:
\begin{equation}\small
(\pi_n,\, \pi_b,\, \pi_c) = \mathrm{DAH}({\bf Z}),
\end{equation}
where $\pi_n$, $\pi_b$, and $\pi_c$ represent the logits for noise, blur, and low-contrast degradations, respectively.  Based on a threshold $\zeta$, degradation factors are generated as:
\begin{equation}\small
d_l = \mathbf{1}\!\big[\sigma(\pi_l) \ge \zeta\big], \quad l \in \{n,b,c\},
\end{equation}
where $\sigma(\cdot)$ denotes the Sigmoid function.  
When $d_l = 0$, the corresponding degradation is absent, and its DRM branch is explicitly suppressed, reducing the risk of over-restoration.

In the DEH, degradation intensity is modeled as a continuous variable in $[0,1]$ using EDL~\cite{sensoy2018evidential}, which enables reliable estimation by jointly predicting both intensity and confidence. 
Specifically, the DEH outputs the parameters of a Beta distribution:
\begin{equation}\small
(\alpha,\, \beta) = \mathrm{DEH}({\bf Z}), \quad
p = \frac{\alpha}{\alpha+\beta}, \quad S = \alpha+\beta,
\end{equation}
where $p$ denotes the degradation intensity, $S$ represents confidence, and $\alpha$ and $\beta$ correspond to the ``evidence'' for existence and nonexistence, respectively. The estimated $p$ and $S$ are concatenated and fed into a fully connected (FC) layer to produce channel modulation parameters:
\begin{equation}\small
\boldsymbol{\gamma},\, \boldsymbol{\bar{\beta}} 
= \mathrm{FC}\!\big(\mathrm{Cat}(\operatorname{logit}(p),\, \log S)\big),
\end{equation}
where $\operatorname{logit}(p) = \log\frac{p}{1-p}$. Finally, intensity-aware modulation is applied to the encoder features to inject the degradation priors:
\begin{equation}\small
{\bf F} = \mathrm{Enc}({\bf I}^d) \otimes (1 + \boldsymbol{\gamma}) + \boldsymbol{\bar{\beta}},
\end{equation}
where $\mathrm{Enc}(\cdot)$ denotes the encoder consisting of multiple convolutional layers and ResBlocks, and $\otimes$ denotes channel-wise multiplication.
Guided by both intensity and confidence, this dynamic modulation establishes a mapping between encoder outputs and degradation severity, enabling the model to adaptively adjust feature distributions according to the restoration requirements of each DRM for more accurate degradation correction.

During training, the DAH is optimized with binary cross-entropy with logits loss:
\begin{equation}\small
\label{eq:bce}
\mathcal{L}_{\mathrm{BCE}}
= \frac{1}{B}\sum_{j=1}^{B}\sum_{l\in\{n,b,c\}}\big(\mathrm{softplus}(\pi^{(j)}_l) - y^{(j)}_l\, \pi^{(j)}_l\big),
\end{equation}
where $\mathrm{softplus}(x) = \log(1 + e^x)$, $y_l$ denotes the degradation label and $B$ is the batch size.
For the DEH, the degradation intensity label is defined based on the Structural Similarity Index (SSIM)~\cite{wang2004image}:
\begin{equation}\small
y^d = 1 - \mathrm{SSIM}({\bf I}^d,\, {\bf I}^c),
\end{equation}
where ${\bf I}^c$ is the clean reference image corresponding to ${\bf I}^d$.  Based on this, the Beta–EDL Loss~\cite{sensoy2018evidential} is formulated as:
\begin{equation}\small
\begin{aligned}
\label{eq:edl}
\mathcal{L}_{\mathrm{EDL}}
=&~\underbrace{-\Big[(\alpha{-}1)\ln y^d + (\beta{-}1)\ln(1{-}y^d) - \ln B(\alpha,\beta)\Big]}_{\mathcal{L}_{\mathrm{NLL}}}\\
&+ \tau \cdot \underbrace{\mathrm{KL}\!\big(\mathrm{Beta}(\alpha,\beta)\,\|\,\mathrm{Beta}(1,1)\big)}_{\mathcal{L}_{\mathrm{KL}}},
\end{aligned}
\end{equation}
where $B(\alpha,\beta)$ denotes the Beta function, $\mathrm{Beta}(\cdot,\cdot)$ the Beta distribution, and $\tau$ the annealing coefficient.
The negative log-likelihood term $\mathcal{L}_{\mathrm{NLL}}$ fits the predicted Beta distribution to the target degradation intensity $y^d$, while the KL regularization term $\mathcal{L}_{\mathrm{KL}}$ penalizes overconfidence, ensuring accurate intensity estimation and reliable confidence calibration.  The overall training objective for DENet is defined as:
\begin{equation}
\mathcal{L}_{\mathrm{DENet}} = \mathcal{L}_{\mathrm{BCE}} + \mathcal{L}_{\mathrm{EDL}}.
\end{equation}

By combining degradation-aware and confidence-driven intensity modeling, DENet provides explicit and reliable degradation priors for subsequent DRM branches, enabling the model to adaptively perceive degradation variations and dynamically regulate each branch’s restoration process.

\subsection{Degradation Residual Modules}
\label{sec:DRM}
Different types of degradation within compound degradations exhibit significant differences in their physical mechanisms and feature distributions. A single shared network struggles to handle multiple degradations simultaneously, often suffering from cross-degradation interference and parameter competition during optimization. 
To address this issue, we explicitly decompose compound degradations into multiple independent sub-processes and design a DRM for each degradation type to achieve decoupled modeling. As illustrated in Figure~\ref{fig:model}(II), three DRMs are constructed for noise, blur, and low-contrast degradations, respectively. These DRMs share identical architectures but maintain independent parameters. Each DRM is built upon a RestormerBlock~\cite{zamir2022restormer}, which learns the residual components corresponding to its specific degradation type. 

Let $\mathbf{F}^{pre}$ denote the input feature (which can be the encoder output $\mathbf{F}$ or the output of another DRM). The feature for the $l$-th degradation type is updated as:
\begin{equation}\small
\mathbf{F}_l = \mathbf{F}^{pre} - d_l \cdot \mathrm{RB}_l(\mathbf{F}^{pre}),
\end{equation}
where $\mathrm{RB}_l$ denotes the RestormerBlock designed for degradation type $l$, and $d_l \in \{0,1\}$ is the degradation factor provided by DENet. When $d_l=1$, the corresponding DRM is activated for targeted restoration, whereas when $d_l=0$, the branch is explicitly disabled to prevent over-restoration in non-degraded cases.  This divide-and-conquer residual modeling strategy allows each DRM to learn degradation-specific compensation within its own parameter space, effectively mitigating cross-degradation gradient conflicts and parameter competition. Moreover, guided by the degradation type and intensity priors from DENet, each DRM achieves fine-grained, degradation-aware adaptive restoration, significantly improving modeling precision and stability under complex compound degradation scenarios.

\subsection{SE-Guided Restoration Order Selection}
\label{sec:SEROS}
Compound degradations exhibit strong nonlinear dependencies and couplings, where the removal order of different degradations plays a crucial role in the final restoration quality. 
An inappropriate restoration order may alter the degradation distribution of subsequent stages, leading to error accumulation and loss of structural details. 
For instance, performing deblurring under severe noise conditions may cause noise to corrupt blur kernel estimation, resulting in pseudo textures, whereas applying denoising first under heavy blur may weaken edge consistency. 
To address this issue, the proposed SE-ROS aims to explicitly model the order dependency among degradation removal processes and adaptively select the optimal restoration paths through the 2D-SE minimization criterion, thereby enabling stable and high-fidelity restoration under compound degradations.

As illustrated in Figure~\ref{fig:model}(III), we first enumerate all possible permutations of the three degradation types to construct a set of candidate features:
\begin{equation} \small
\mathcal{F} = \{\mathbf F_{cbn},\ \mathbf F_{cnb},\ \mathbf F_{bcn},\  \mathbf F_{bnc},\ \mathbf F_{ncb},\ \mathbf F_{nbc}\}.
\end{equation}
We then model $\mathcal{F}$ as a fully connected similarity graph $G=(\mathcal V, \mathcal E)$, where each vertex represents a candidate feature and each edge weight is defined by the non-negative cosine similarity:
\begin{equation}\small
w_{ij} = 
\max\!\big(\mathrm{Cos}(\mathbf F_i, \mathbf F_j),\, 0\big),
\quad \text{where } \mathbf F_i, \mathbf F_j \in \mathcal{F} \text{ and } i \neq j.
\end{equation}
Where $\mathrm{Cos}(\cdot,\cdot)$ denotes the cosine similarity operator, and $\max(\cdot,\cdot)$ denotes the maximum operation.  Let $\mathcal{P}=\{\mathcal C_1,\dots,\mathcal C_k\}$ denote a partition of $G$. 
The 2D-SE~\cite{li2016structural} is then defined as:
\begin{equation}\small
\label{eq:2dse}
\mathcal{H}^{2}(\mathcal{P})
= - \sum_{\mathcal C \in \mathcal{P}} \!\left( \frac{g_{\mathcal C}}{v_G} \log \frac{v_\mathcal C}{v_G}
+ \sum_{x \in \mathcal C} \frac{o_x}{v_G} \log \frac{o_x}{v_\mathcal C} \right),
\end{equation}
where $o_x$ denotes the degree of vertex $x$, $v_G$ is the graph volume (the sum of all vertex degrees), $g_\mathcal C$ is the cut weight of part $\mathcal C$, and $v_\mathcal C$ is the volume of $\mathcal C$.  
By minimizing $\mathcal{H}^{2}(\mathcal{P})$, SE-ROS adaptively identifies the most informative restoration orders, effectively capturing order dependencies in an interpretable information-theoretic manner.

We employ a near-linear-time 2D-SE minimization algorithm~\cite{xian2025community} to obtain the optimal second-order coding tree and corresponding partition $\mathcal P=\{\mathcal C_1,\dots,\mathcal C_k\}$, ensuring high intra-cluster similarity and maximal inter-cluster separability. 
The 2D-SE contribution of each vertex is defined as the increase in SE incurred when the vertex is removed from its current part $\mathcal C_j$:
\begin{equation}\small
\label{Eq:SE_con}
\begin{aligned} 
\mathcal P^\prime&=\{\mathcal C_1,\dots,\mathcal C_j\!\setminus\!\{x\},\dots,\mathcal C_k,\{x\}\},\\ 
\Delta H_x&=\mathcal H^2(\mathcal P')-\mathcal H^2(\mathcal P)\\ 
&= -\frac{g_{\mathcal C_j^\prime}}{v_G} \log \frac{v_{\mathcal C_j^\prime}}{v_G} 
+ \frac{g_{\mathcal C_j}}{v_G} \log \frac{v_{\mathcal C_j}}{v_G}
- \frac{o_x}{v_G} \log \frac{v_{\mathcal C_j}}{v_G} 
- \frac{v_{\mathcal C_j^\prime}}{v_G} \log \frac{v_{\mathcal C_j}}{v_{\mathcal C_j^\prime}}, 
\end{aligned} 
\end{equation}
where $\mathcal C_j^\prime = \mathcal C_j \setminus \{x\}$. A larger $\Delta H_x$ indicates that vertex $x$ (i.e., the candidate feature $\mathbf F_x$) contributes more to maintaining the global information order. Therefore, we select the vertex with the highest 2D-SE contribution within each part and perform weighted aggregation as follows:
\begin{equation} \small
\label{eq:H}
\begin{aligned} 
\mathcal{F}^s &= \big[\arg\max_{x\in \mathcal C_1}\Delta \mathcal H_x,\dots,\arg\max_{x\in \mathcal C_k}\Delta \mathcal H_x\big]^\top,\\ 
\mathbf s&= \mathrm{softmax}\!\Big(\big[\max_{x\in \mathcal C_1}\Delta \mathcal H_x,\dots,\max_{x\in \mathcal C_k}\Delta \mathcal H_x\big]^\top\Big),\\ 
\mathbf F'&=\sum_{i=1}^{k}\, \mathbf s_i\,\mathbf F_i,\quad \mathbf F_i\in \mathcal{F}^s. 
\end{aligned} 
\end{equation} 
Here, $\mathrm{softmax}(\cdot)$ denotes the Softmax function, $\mathbf s$ represents the aggregation weights, and $\mathbf F^\prime$ denotes the final aggregated feature. 
This 2D-SE-based feature selection and weighted aggregation preserve the complementary information among different restoration paths while suppressing redundancy and artifacts. 
Finally, the decoder generates the high-quality restored image:
\begin{equation}\small
\mathbf I^r = \mathrm{Dec}(\mathbf F^\prime), \quad
\mathcal{L}_{\ell_1} = \|\mathbf I^r - \mathbf I^c\|_1,
\end{equation}
where $\mathrm{Dec}(\cdot)$ denotes the main decoder composed of multiple convolutional layers and ResBlocks.  

In this process, SE-ROS transforms the restoration path selection into an information-theoretic optimization problem. By minimizing the 2D-SE, SE-ROS jointly performs restoration order modeling and feature aggregation under a unified objective, enabling adaptive path selection guided by degradation-structure information. This mechanism operates without additional supervision, dynamically selecting the most informative candidate features based on degradation characteristics while suppressing redundancy and false components. Consequently, it significantly reduces multi-stage error propagation and artifact amplification. Moreover, the computation of the optimal second-order coding tree and vertex contributions ($\Delta \mathcal H_x$) provides clear interpretability and strong generalization stability for the restoration path selection.

\begin{table*}[t]
    \centering
    \setlength{\tabcolsep}{2.8pt}
    \caption{Quantitative comparison on the  HM-TIR and Night-TIR datasets. The best and second-best performances for each metric are highlighted with \colorbox[HTML]{ffc7ce}{Red} and \colorbox[HTML]{d9e7f4}{Blue} backgrounds, respectively.}
    \small
    \label{tab:QC_HN}  
    \begin{threeparttable}
        \vspace{-2mm}
        \begin{tabularx}{\linewidth}{l|cc|cc|cc|cc|cc|cc|cc|cc}
        \toprule
        Dataset&\multicolumn{8}{c|}{HM-TIR}&\multicolumn{8}{c}{Night-TIR}\\
        \midrule
         Degradation&\multicolumn{2}{c|}{Single}&\multicolumn{2}{c|}{Double}& \multicolumn{2}{c|}{Triple}&\multicolumn{2}{c|}{Avg.}&
        \multicolumn{2}{c|}{Single}&\multicolumn{2}{c|}{Double}& \multicolumn{2}{c|}{Triple}&\multicolumn{2}{c}{Avg.}\\
        \midrule
         Metric&PSNR&SSIM&PSNR&SSIM&PSNR&SSIM&PSNR&SSIM&PSNR&SSIM&PSNR&SSIM&PSNR&SSIM&PSNR&SSIM\\
        \midrule
        WFAF~\cite{munch2009stripe} &\footnotesize 22.312& \footnotesize 0.757&\footnotesize 17.849& \footnotesize 0.560&\footnotesize 16.422&\footnotesize 0.372&\footnotesize 18.861&\footnotesize 0.563&\footnotesize 21.960 &\footnotesize 0.766 &\footnotesize 17.294 &\footnotesize 0.529&\footnotesize 16.063 &\footnotesize 0.368 &\footnotesize 18.439 &\footnotesize 0.555\\
        LRSID~\cite{chang2016remote}&\footnotesize 23.382& \footnotesize 0.787&\footnotesize 18.065& \footnotesize 0.609&\footnotesize 16.561&\footnotesize 0.444&\footnotesize 19.336&\footnotesize 0.613&\footnotesize 22.283 &\footnotesize 0.783 &\footnotesize 17.420 &\footnotesize 0.571&\footnotesize 16.160 &\footnotesize 0.426 &\footnotesize 18.621 &\footnotesize 0.593\\
        TSIRE~\cite{pang2023infrared}&\footnotesize 19.068& \footnotesize 0.649&\footnotesize 18.550& \footnotesize 0.484&\footnotesize 16.469&\footnotesize 0.201&\footnotesize 18.028&\footnotesize 0.445&\footnotesize 18.185 &\footnotesize 0.661 &\footnotesize 16.997 &\footnotesize 0.447&\footnotesize 15.399 &\footnotesize 0.210 &\footnotesize 16.860 &\footnotesize 0.439\\
        DiffUIR~\cite{zheng2024selective}&\footnotesize 22.168& \footnotesize 0.756&\footnotesize 17.683& \footnotesize 0.599&\footnotesize 16.149&\footnotesize 0.336&\footnotesize 18.667&\footnotesize 0.564&\footnotesize 21.392 &\footnotesize 0.801 &\footnotesize 17.213 &\footnotesize 0.571&\footnotesize 15.923 &\footnotesize 0.329 &\footnotesize 18.176 &\footnotesize 0.567\\
        DACLIP~\cite{DACLIP2024}&\footnotesize 22.130& \footnotesize 0.733&\footnotesize 17.819& \footnotesize 0.551&\footnotesize 16.480&\footnotesize 0.399&\footnotesize 18.810&\footnotesize 0.561&\footnotesize 21.279 &\footnotesize  0.735 &\footnotesize 17.520 &\footnotesize 0.541&\footnotesize 16.085 &\footnotesize 0.391 &\footnotesize 18.295 &\footnotesize 0.556\\
        PPFN~\cite{liu2025enhancing}&\footnotesize \cellcolor[HTML]{d9e7f4}{\strut 31.121}& \footnotesize \cellcolor[HTML]{d9e7f4}{\strut 0.937}&\footnotesize \cellcolor[HTML]{d9e7f4}{\strut 23.761}& \footnotesize \cellcolor[HTML]{d9e7f4}{\strut 0.825}&\footnotesize \cellcolor[HTML]{d9e7f4}{\strut 18.201}&\footnotesize \cellcolor[HTML]{d9e7f4}{\strut 0.724}&\footnotesize \cellcolor[HTML]{d9e7f4}{\strut 24.361}&\footnotesize \cellcolor[HTML]{d9e7f4}{0.829}&\footnotesize \cellcolor[HTML]{ffc7ce}{27.016}& \footnotesize \cellcolor[HTML]{d9e7f4}{0.919}&\footnotesize \cellcolor[HTML]{d9e7f4}{21.010}& \footnotesize \cellcolor[HTML]{d9e7f4}{0.794}&\footnotesize \cellcolor[HTML]{d9e7f4}{17.762}&\footnotesize \cellcolor[HTML]{d9e7f4}{0.704} &\footnotesize \cellcolor[HTML]{d9e7f4}{21.929} &\footnotesize \cellcolor[HTML]{d9e7f4}{0.806}\\
        SEGD (ours)&\footnotesize \cellcolor[HTML]{ffc7ce}{\strut 31.601}& \footnotesize  \cellcolor[HTML]{ffc7ce}{\strut 0.951}&\footnotesize  \cellcolor[HTML]{ffc7ce}{\strut 26.342}& \footnotesize  \cellcolor[HTML]{ffc7ce}{\strut 0.872}&\footnotesize  \cellcolor[HTML]{ffc7ce}{\strut 20.310}&\footnotesize  \cellcolor[HTML]{ffc7ce}{\strut 0.761}&\footnotesize  \cellcolor[HTML]{ffc7ce}{\strut 26.085}&\footnotesize  \cellcolor[HTML]{ffc7ce}{\strut 0.862}&\footnotesize  \cellcolor[HTML]{d9e7f4}{\strut 26.789} &\footnotesize  \cellcolor[HTML]{ffc7ce}{\strut 0.922} &\footnotesize  \cellcolor[HTML]{ffc7ce}{\strut 22.323} &\footnotesize  \cellcolor[HTML]{ffc7ce}{\strut 0.842}&\footnotesize  \cellcolor[HTML]{ffc7ce}{\strut 18.312} &\footnotesize  \cellcolor[HTML]{ffc7ce}{\strut 0.737} &\footnotesize  \cellcolor[HTML]{ffc7ce}{\strut 22.475} &\footnotesize  \cellcolor[HTML]{ffc7ce}{\strut 0.834}\\
        \bottomrule
        \end{tabularx}
    \end{threeparttable}\vspace{-1mm}
\end{table*}

\section{Experiments}
\label{sec:experiments}

\subsection{Experimental Setups}
\textbf{Datasets}.
We train and evaluate SEGD on the HM-TIR~\cite{liu2025enhancing} dataset, which contains 1{,}503 TIR images randomly divided into training and test sets with an 8:2 ratio. The degraded inputs are synthesized strictly following the degradation protocol of PPFN~\cite{liu2025enhancing}. The test set includes 20\% single-degradation, 30\% double-degradation, and 50\% triple-degradation cases. In addition, we construct a night-scene benchmark, Night-TIR (647 images), and generate the corresponding degraded versions using the same procedure as HM-TIR. Details of Night-TIR’s image sources and statistics are provided in the supplementary material.
To further evaluate generalization under real-world conditions, we randomly sample 300 TIR images from AWMM~\cite{li2026all} as a real-scene test set, without applying any synthetic degradations, to assess robustness under natural distributions and cross-scene domain shifts.

\textbf{Evaluation metrics}.
For HM-TIR and Night-TIR, we report Peak Signal-to-Noise Ratio (PSNR) and Structural Similarity Index (SSIM)~\cite{wang2004image}. For AWMM, we adopt three commonly used no-reference image quality assessment metrics: NIMA~\cite{talebi2018nima} and MUSIQ~\cite{ke2021musiq} (higher indicates better perceptual quality), and NIQE~\cite{mittal2012making} (lower indicates better quality).

\textbf{Implementation Details}.
All experiments are implemented in PyTorch and conducted on an NVIDIA RTX 4090 GPU with 48 GB of VRAM.
We employ the AdamW optimizer with a batch size of 6. Training patches of size $256 \times 256$ are obtained through sliding-window cropping. The initial learning rate is set to $5 \times 10^{-5}$ with a cosine-annealing schedule, and the model is trained for 100 epochs. During inference, the degradation threshold is fixed at $\zeta = 0.45$. Additional architectural and implementation details are provided in the supplementary material.

\textbf{Comparison Methods}.
We compare SEGD with four TIR-oriented enhancement methods—WFAF~\cite{munch2009stripe}, LRSID~\cite{chang2016remote}, TSIRIE~\cite{pang2023infrared}, and PPFN~\cite{liu2025enhancing}—as well as two all-in-one restoration models designed for visible images, DACLIP~\cite{DACLIP2024} and DiffUIR~\cite{zheng2024selective}, for comprehensive evaluation and analysis. All methods follow the experimental protocol established by PPFN~\cite{liu2025enhancing}.

\begin{table}[t!]
    \centering
    \setlength{\tabcolsep}{2pt} 
    \caption{Parameters and inference time comparison on the HM-TIR dataset. The best and second-best results for each metric are highlighted with \colorbox[HTML]{ffc7ce}{Red} and \colorbox[HTML]{d9e7f4}{Blue} backgrounds, respectively.}
    \label{tab:efficiency}
    \footnotesize
 \begin{threeparttable}
        \vspace{-2mm}
        \begin{tabularx}{\linewidth}{l|*{7}{>{\centering\arraybackslash}c}}
            \toprule
             Methods& WFAF & LRSID & TSIRE & DiffUIR & DACLIP & PPFN & SEGD\\
            \midrule 
             Params (M)& -- & -- &\cellcolor[HTML]{d9e7f4}{\strut 2.52} &12.41 &233.14 & 26.60 &\cellcolor[HTML]{ffc7ce}{\strut 2.268} \\
             Time (S)& 0.26 & 4.58 &\cellcolor[HTML]{ffc7ce}{\strut 0.01} &0.43 &15.32 & 0.65 & \cellcolor[HTML]{d9e7f4}{\strut 0.19}\\
            \bottomrule
        \end{tabularx}
    \end{threeparttable}\vspace{-4mm}
\end{table}

\begin{figure*}[t]
    \centering
    \includegraphics[width=0.95\linewidth, trim={0cm 0cm 0cm 0cm}, clip]{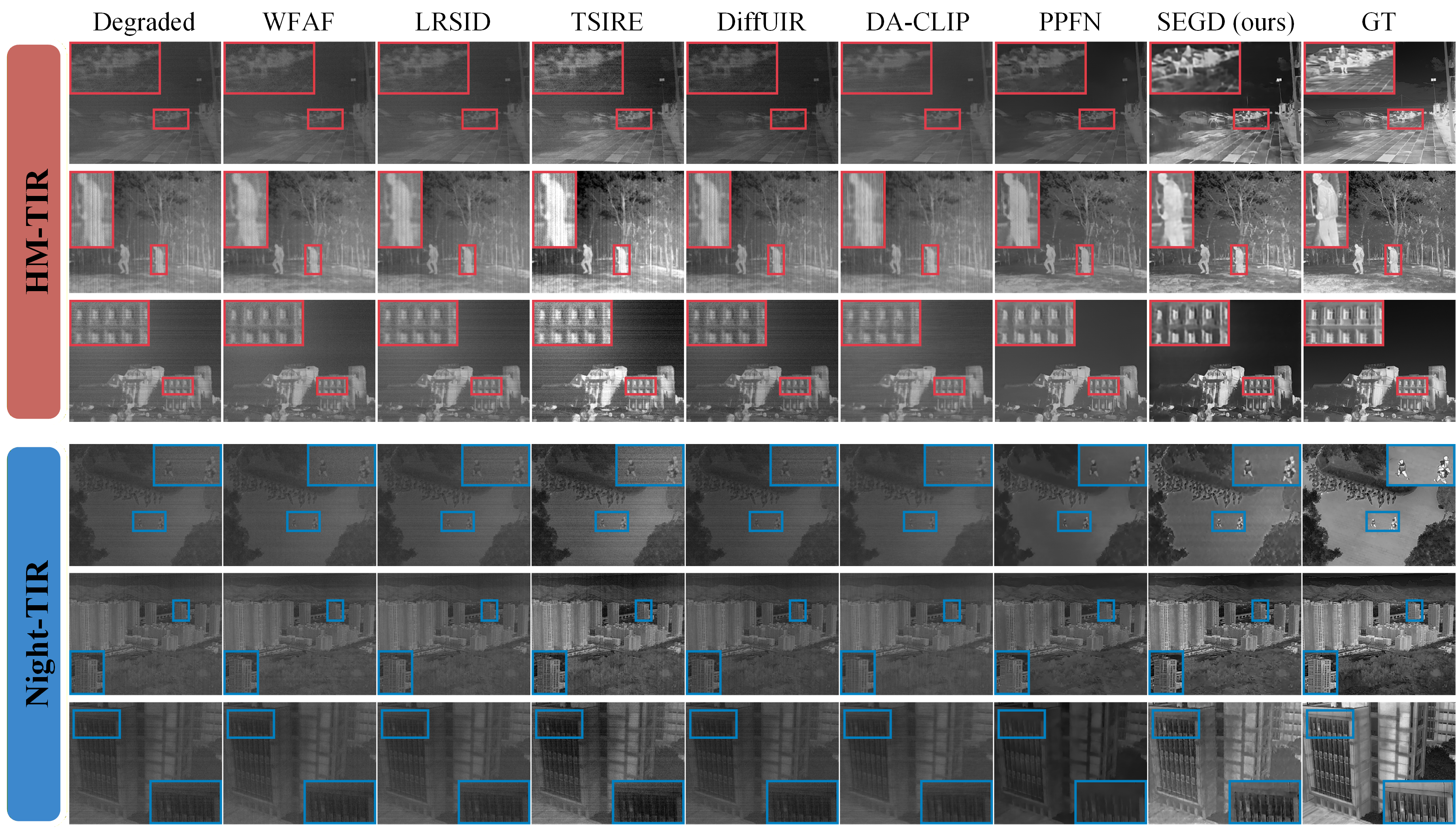}
    \vspace{-2mm}
    \caption{Qualitative comparison under triple compound degradations on the HM-TIR and Night-TIR datasets. Due to space constraints, qualitative results for single- and double-degradation settings are provided in the supplementary material.}
    \label{fig:QC_HN}\vspace{-3mm}
\end{figure*}
\subsection{Results on Compound Degradation TIR}
We evaluate the proposed SEGD on HM-TIR and Night-TIR under single-, double-, and triple-degradation settings.
As shown in Table~\ref{tab:QC_HN}, SEGD achieves the best PSNR and SSIM on both datasets—with average gains of 4.90\% and 3.73\%, respectively—and its advantage becomes more pronounced under compound degradations. 
Specifically, WFAF and LRSID are tailored to specific degradations and drop markedly when degradations are compounded. DACLIP and DiffUIR generalize poorly to TIR due to disparities in image formation mechanisms and degradation distributions between visible and thermal domains. TSIRIE employs a single shared backbone to handle all degradations, which leads to task competition and gradient conflicts, resulting in weaker performance. PPFN shows better stability by leveraging degradation context and selective progressive training, but it still depends on a shared backbone and ignores degradation intensity variations. In contrast, SEGD jointly perceives degradation type and intensity, explicitly models different degradations in a divide-and-conquer manner with coordinated optimization, and adaptively selects optimal restoration paths via the 2D-SE criterion, thereby achieving higher PSNR and SSIM. Qualitative results (Figure~\ref{fig:QC_HN}) further corroborate these findings: most methods—except PPFN—exhibit artifacts and residual background noise, while PPFN still suffers from detail loss and insufficient contrast. SEGD balances contrast enhancement, detail fidelity, and artifact suppression under complex compound degradations, producing cleaner and more structurally faithful TIR images.
Notably, compared with PPFN, SEGD attains higher inference efficiency with fewer parameters (see Table~\ref{tab:efficiency}).

\subsection{Results on Real-world TIR}
\begin{figure*}[t]
    \centering
    \includegraphics[width=0.93\linewidth, trim={0cm 0cm 0cm 0cm}, clip]{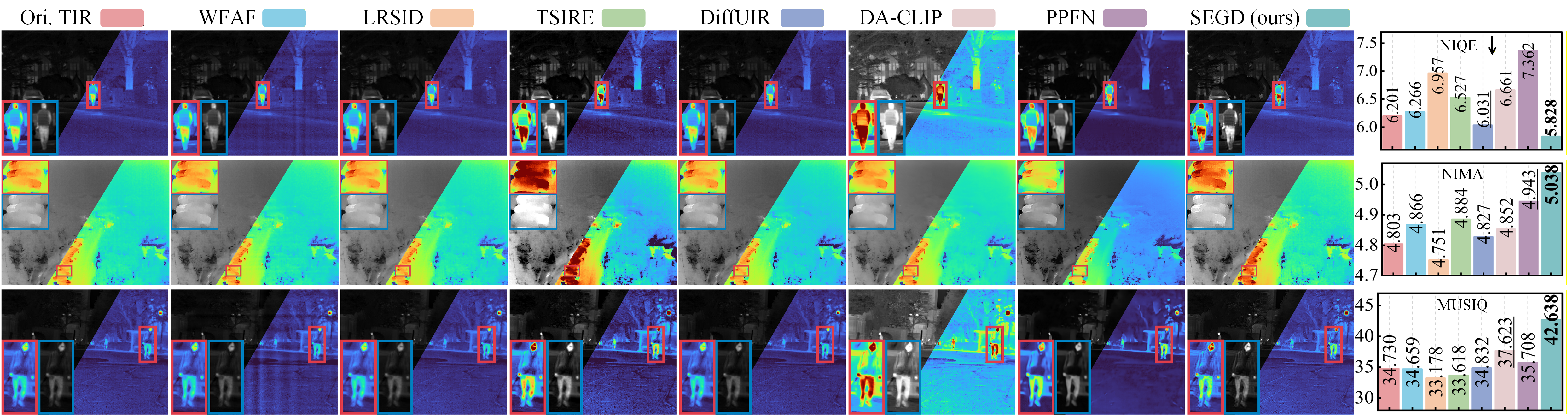}
    \vspace{-2mm}
    \caption{Quantitative and qualitative comparisons on the AWMM dataset. }
    \label{fig:QA_AWMM}\vspace{-3mm}
\end{figure*}

We further evaluate the generalization and perceptual quality of SEGD on the AWMM dataset.
As shown in Figure~\ref{fig:QA_AWMM}, SEGD simultaneously preserves sharpness, suppresses noise, and maintains thermal semantic consistency. It produces clearer boundaries and repetitive structures in high-frequency regions (e.g., pedestrian contours and ground textures), effectively reduces background noise, and avoids over-enhancement, yielding a more natural overall appearance. These advantages are consistent with the statistics on the right—SEGD achieves the lowest NIQE and the highest NIMA/MUSIQ, indicating improvements aligned with human perception in both global contrast and local detail. By comparison, WFAF, LRSID, TSIRIE, and PPFN struggle to perceive latent degradation types and intensities; in strong-noise or weak-texture regions, they tend to over-smooth or over-restore, leading to blurred object boundaries and residual background granularity. Visible-domain methods (DACLIP and DiffUIR) perform inconsistently on TIR images, often causing local saturation or discontinuous thermal gradients. Overall, SEGD surpasses competing methods in both perceptual quality and generalization.
\begin{table}[b]
    \vspace{-3mm}
    \centering
    \setlength{\tabcolsep}{3.5pt} 
    \caption{Ablation studies on the DAH and DEH in DENet. The best and second-best performances for each metric are highlighted with \colorbox[HTML]{ffc7ce}{Red} and \colorbox[HTML]{d9e7f4}{Blue} backgrounds, respectively.}
    \label{tab:Ablation_DENet}
    \small
 \begin{threeparttable}
        \vspace{-2mm}
        \begin{tabularx}{\linewidth}{c*{6}{>{\centering\arraybackslash}c}}
            \toprule
             &\multirow{2}{*}{DAH} &\multirow{2}{*}{DEH} &\multirow{2}{*}{\makecell{Degradation\\Types (GT)}}&\multirow{2}{*}{\makecell{Confidence \\ ($S$)}} & \multicolumn{2}{c}{Avg. Metrics}  \\
             \cmidrule{6-7}
            &&&&&PSNR&SSIM\\
            \midrule 
            1&\ding{55}&\ding{55}&\ding{55}&\ding{55}&22.728&0.823\\
            2&\ding{55}&\ding{51}&\ding{55}&\ding{51}&24.263&0.841\\
            3&\ding{51}&\ding{55}&\ding{55}&\ding{55}&25.066&0.850\\
            4&\ding{51}&\ding{51}&\ding{55}&\ding{55}&25.830&0.857\\
            5&\ding{51}&\ding{51}&\ding{55}&\ding{51}&\cellcolor[HTML]{d9e7f4}{\strut 26.085}&\cellcolor[HTML]{d9e7f4}{\strut 0.862}\\
            6&\ding{55}&\ding{51}&\ding{51}&\ding{51}&\cellcolor[HTML]{ffc7ce}{\strut26.199}&\cellcolor[HTML]{ffc7ce}{\strut 0.863}\\
            \bottomrule
        \end{tabularx}
    \end{threeparttable}
\end{table}

\subsection{Ablation Studies}
This section presents ablation results on HM-TIR, quantifying the individual contributions of DENet, DRMs, and SE-ROS to verify the effectiveness and necessity of each component. Additional ablation studies and analyses are provided in the supplementary material.

\begin{figure}[h]
    \centering
    \includegraphics[width=0.93\linewidth, trim={0cm 0.2cm 0cm 0cm}, clip]{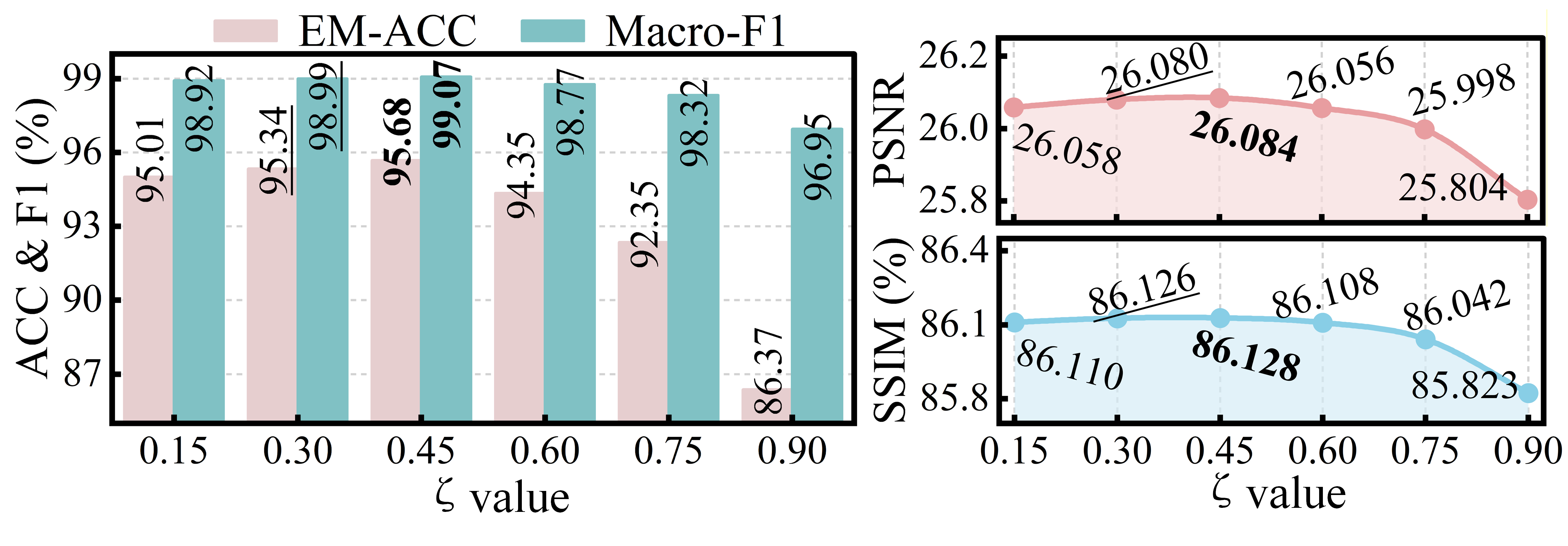}
    \vspace{-2mm}
    \caption{The effect of varying thresholds ($\zeta$) on performance. }
    \label{fig:DNE}\vspace{-3mm}
\end{figure}

\textbf{Effectiveness of DENet}.
As shown in Table~\ref{tab:Ablation_DENet}, removing either DAH or DEH degrades performance, with the largest drop when both are removed. Using only DEH’s intensity \(P\) without confidence \(S\) causes a small decline, confirming that \(S\) suppresses unreliable estimates; replacing DAH with ground-truth types yields only marginal gains and is impractical in real-world scenarios. 
Additionally, DENet is largely insensitive to the threshold \(\zeta\) (Figure~\ref{fig:DNE}), except at excessively high values (e.g., \(0.9\)).

\begin{figure}[h]
    \vspace{-3mm}
    \centering
    \includegraphics[width=0.93\linewidth, trim={0cm 0cm 0cm 0cm}, clip]{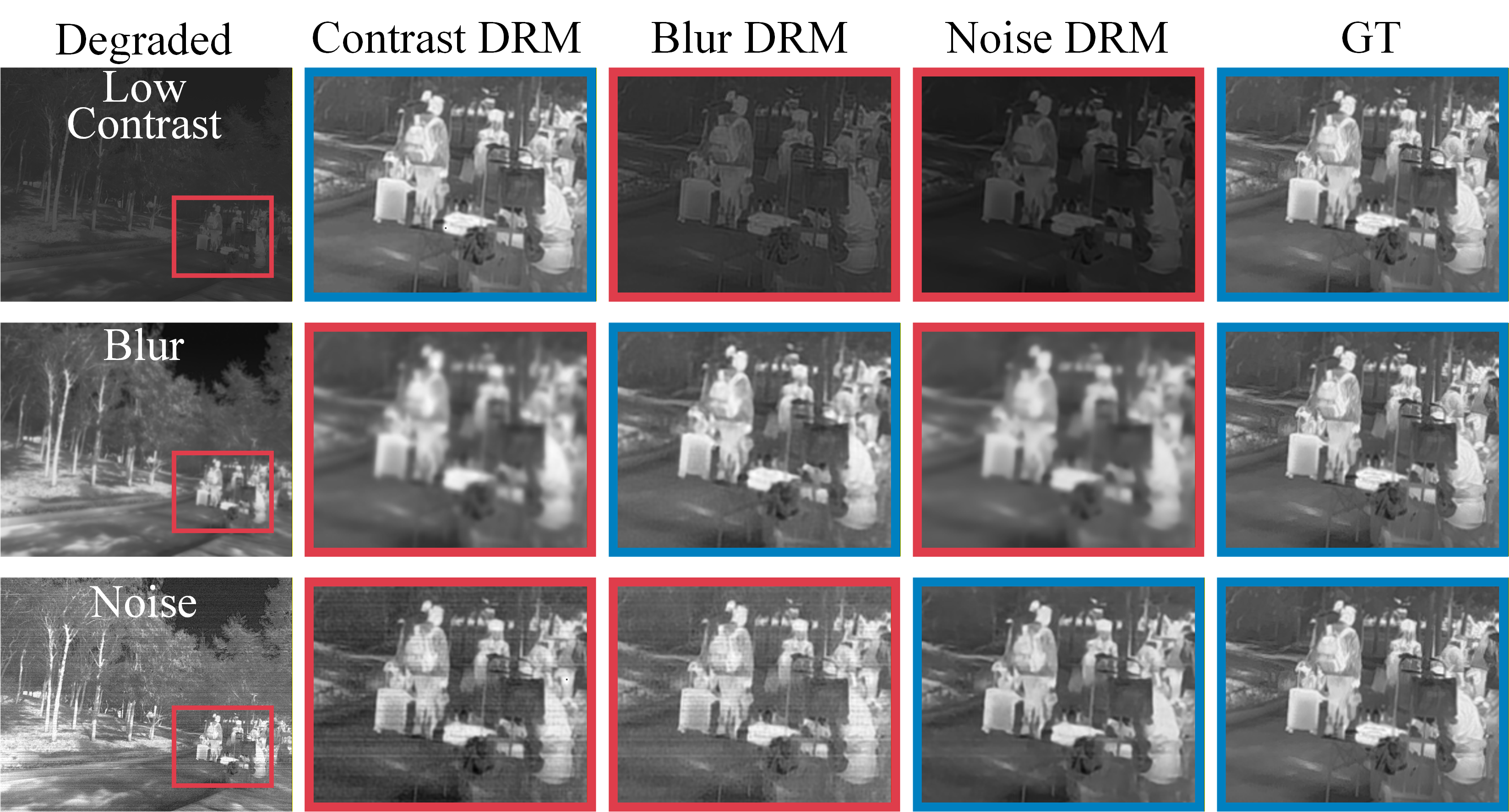}
    \vspace{-1mm}
    \caption{The efficacy of each DRM toward specific degradation.}
    \label{fig:DRM}\vspace{-2mm}
\end{figure}
\textbf{Effectiveness of DRMs}.
Figure~\ref{fig:DRM} shows that each DRM is highly specialized: it yields substantial gains on its target degradation but provides little or no benefit for others, e.g., the contrast DRM improves low-contrast cases but not noise or blur; the noise and blur DRMs exhibit similar behavior. When combined, the three modules complement each other, jointly covering multiple degradations and robustly handling complex compound cases. Notably, removing the DRMs renders both DENet and SE-ROS ineffective, leading to a substantial performance drop---PSNR and SSIM decrease by 6.045~dB and 0.091, respectively.

\begin{figure}[t]
    \centering
    \includegraphics[width=0.95\linewidth, trim={0.2cm 0.2cm 0.2cm 0cm}, clip]{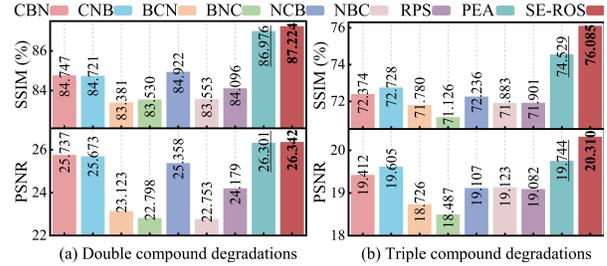}
    \vspace{-1mm}
    \caption{Results on different degradation restoration strategies.}
    \label{fig:SEROS}\vspace{-4mm}
\end{figure}

\textbf{Effectiveness of SE-ROS}.
We compare SE-ROS with three strategies: Fixed Restoration Order (e.g., contrast enhancement $\rightarrow$ deblurring $\rightarrow$ denoising, denoted CBN), Random Path Sampling (RPS), and Pathwise Equal Averaging (PEA). As shown in Figure~\ref{fig:SEROS}, SE-ROS achieves the highest PSNR and SSIM in both double- and triple-compound settings. Fixed orders cannot accommodate the nonlinear coupling among degradations and are prone to error propagation, thereby limiting performance. The stochastic nature of RPS prevents consistent selection of the optimal path for each image, leading to unstable results. In the double-compound case, PEA performs on par with SE-ROS because the two candidate paths contribute equally to the 2D-SE of feature graph, rendering SE-ROS effectively equivalent to PEA. In contrast, under triple-compound degradations, averaging all paths in PEA dilutes information from superior paths and degrades performance.

\section{Conclusion}
\label{sec:conclusion}
We present SEGD, a Structural Entropy–Guided Decoupled framework for infrared enhancement. 
SEGD employs DRMs to model distinct degradation types in independent parameter spaces, decoupling degradations while enabling joint optimization, fundamentally mitigating cross-degradation gradient interference and parameter competition. Furthermore, a DENet is developed to estimate degradation type and strength as priors that drive DRMs to modulate restoration dynamically.  To address nonlinear coupling in compound degradations, we introduce SE-ROS, which adaptively selects image-specific restoration paths via 2D-SE, unifying order selection and feature aggregation. We also construct Night-TIR, a nighttime TIR benchmark that provides a challenging testbed. Experiments demonstrate that SEGD consistently outperforms strong baselines across diverse compound degradation settings while achieving higher efficiency with a smaller model.

\clearpage
\section*{Acknowledgements}
This work was supported in part by the National Natural Science Foundation of China(62571222, 62276120,
62576163, 62161015), and the Yunnan Fundamental Research Projects (202501AS070123, 202301AV070004).
{
    \small
    \bibliographystyle{ieeenat_fullname}
    \bibliography{main}
}
\clearpage
\clearpage
\setcounter{page}{1}
\maketitlesupplementary

\section{Preliminaries}
\label{sec:Prel}
In this section, we summarize the concepts and definitions related to the background of our work, including thermal infrared (TIR) image enhancement, Evidential deep learning, and Structural entropy.

\subsection{TIR Enhancement}
TIR degradations primarily manifest as low contrast, blur, and noise. These effects arise from (i) insufficient target–background temperature differences and compression introduced by radiometric calibration; (ii) point-spread function (PSF) broadening due to defocus, diffraction, turbulence, or motion; and (iii) fixed-pattern noise (FPN) together with readout/photon noise.
Let $I^c$ denote the clean reference image. The observed degraded image $I^d$ can be modeled as
\begin{equation}
\label{eq:tir-forward}
I^d = (\mathcal{N}_s \circ \mathcal{N}_o) \circ (\mathcal{K} \ast \text{C}(I^{c})) + \mathcal{N}_r,
\end{equation}
where $\circ$ denotes  composition operator and $*$ denotes convolution.
$\text{C}(\cdot)$ is a contrast-compression operator and $\mathcal{K}$ is the blur kernel/PSF. 
$\mathcal{N}_o$ and $\mathcal{N}_s$ represent optics-related and stripe-related FPN, respectively, and $\mathcal{N}_r$ denotes readout/photon noise. 
Note that a single image may contain one or multiple unknown degradations, added in a random order.
Given a degraded TIR image $I^d$, TIR enhancement seeks to learn a mapping $f_{\theta}:\, I^d \mapsto I^r$ such that $I^r \approx I^c$.

\subsection{Evidential Deep Learning}
Evidential Deep Learning (EDL)~\cite{sensoy2018evidential} interprets network outputs as Dirichlet evidence parameters in subjective logic, thereby modeling prediction and uncertainty jointly while discouraging unwarranted overconfidence.
For continuous targets on the interval $[0,1]$, we can employ the binary special case of the Dirichlet—the Beta distribution.
A Beta-evidence head outputs nonnegative evidence $(\alpha,\beta)$ for the presence and absence hypotheses, respectively. 
The mean $p=\frac{\alpha}{\alpha+\beta}$ serves as the predicted value while the total evidence $S=\alpha+\beta$ quantifies confidence.
Given a target label $y\in[0,1]$, the negative log-likelihood is minimized under the Beta model together with a KL regularizer that pulls uncertain posteriors toward the uniform prior:
\begin{equation}
\small
\begin{aligned}
\label{eq:edl}
    \mathcal{L}_{EDL} =&  \underbrace{-\Big[(\alpha{-}1)\ln y + (\beta{-}1)\ln(1{-}y) - \ln B(\alpha,\beta)\Big]}_{L_{NLL}}\\
    &+ \tau \cdot \underbrace{\mathrm{KL}\!\big(\mathrm{Beta}(\alpha,\beta)\,\|\,\mathrm{Beta}(1,1)\big)}_{L_{KL}},
\end{aligned}
\end{equation}
where $B(\alpha,\beta)$ is the Beta function and $\mathrm{KL}(\cdot\Vert\cdot)$ denotes the Kullback–Leibler divergence.  $\mathrm{Beta}(\cdot,\cdot)$ is a Beta distribution, $\mathrm{Beta}(1,1)$ is the uniform prior, and $\tau$ is an annealing coefficient that increases from $0$ to $1$ over training.

\subsection{Structural Entropy}
Structural entropy (SE)~\cite{li2016structural} quantifies the uncertainty and information content of a graph.
Lower SE indicates a more ordered structure.
SE can be defined on coding trees of different heights to measure structural information at different orders.
Given a graph $G(\mathcal{V},\mathcal{E})$, a coding tree $\mathcal{T}$ is a hierarchical partition of the vertex set $\mathcal{V}$ that satisfies~\cite{li2016structural}:
\begin{enumerate}
    \item Each node $x$ in $\mathcal{T}$ is associated with a set $T_{x} \subseteq \mathcal{V}$. 
    For the root node $\lambda$ of $\mathcal{T}$, $T_{\lambda} = \mathcal{V}$. 
    Any leaf node $q$ in $\mathcal{T}$ is associated with a single node in $G$, i.e., $\mathcal{T}_q = \{v\}$, $v \in \mathcal{V}$

    \item For each node $a$ in $\mathcal{T}$, denote all its children as $b_1, \dots, b_k$, then $T_{b_1}, \dots, T_{b_k}$ is a partition of $T_{a}$.

    \item For each node $a$ in $\mathcal{T}$, denote its height as $h(a)$.
    Let $h(\lambda) = 0$ and $h(\bar{a}) = h(a)+1$, where $\bar{a}$ is the parent of $a$. The height of $\mathcal{T}$ , $h(\mathcal{T}) = \max\limits_{a \in \mathcal{T}}{h(a)}$.
\end{enumerate}
The SE of graph $G$ on coding tree $\mathcal{T}$ is defined as:
\begin{equation}
\label{eq:SE}
    \mathcal{H}^{\mathcal{T}}(G) = - \sum_{a \in\mathcal{T},a \neq \lambda  } \frac{g_a}{vol(\lambda)}\log \frac{vol(a)}{vol(\bar{a})},
\end{equation}
where $g_a$ is the summation of the degrees of the cut edges of $\mathcal{T}_a$ (i.e., the weight sum of edges with exactly one endpoint in $\mathcal{T}_a$).
$vol(a)$, $vol(\bar{a})$, and $vol(\lambda)$ represent the volumes, i.e., the sums of node degrees within $\mathcal{T}_a$, $\mathcal{T}_{\bar{a}}$ and $\mathcal{T}_\lambda$, respectively.
The \( d \)-dimensional SE of \( G \), defined as $\mathcal{H}^{(d)}(G) = \min_{\forall \mathcal{T}: h(\mathcal{T}) = d} \{ \mathcal{H}^{\mathcal{T}} (G) \}$, is realized by acquiring an optimal coding tree of height \( d \), in which the disturbance derived from stochastic variation is minimized.

The two-dimensional structural entropy (2D-SE) measures second-order structural information, and minimizing the 2D-SE of a graph reveals the true second-order structure in disordered graphs.
Let $\mathcal{P}={\mathcal{C}1, \mathcal{C}2, \dots, \mathcal{C}k}$ be a partition of $G$, the corresponding 2D-SE is defined as
\begin{equation}
\label{eq:2dse}
        \mathcal{H}^{2}(\mathcal{P})  
        = - \sum_{\mathcal C \in \mathcal{P} } \left( \frac{g_{\mathcal C}}{v_G} \log \frac{v_\mathcal C}{v_G} + \sum_{x \in \mathcal C }  \frac{o_x}{v_G} \log \frac{o_x}{v_\mathcal C} \right),
\end{equation}
where $o_x$ is the degree of vertex $x$ (the sum of weights of edges incident to $x$) and $v_G$ is the graph volume (the sum of all vertex degrees).  $g_{\mathcal{C}}$ and $v_{\mathcal{C}}$ denote the cut weight and the volume of part $\mathcal{C}$, respectively.
Inspired by literature~\cite{xian2025community}, the per-node 2D-SE contribution can be derived in detail as follows:
\begin{equation}
    \small
    \begin{aligned}
    &\mathcal P=\{\mathcal C_1,\dots,\mathcal C_j,\dots,\mathcal C_k\},\\ 
    &\mathcal P^\prime=\{\mathcal C_1,\dots,\mathcal C_j\!\setminus\!\{x\},\dots,\mathcal C_k,\{x\}\},\\ 
    &\Delta H_x =\mathcal H^2(\mathcal P^\prime)-\mathcal H^2(\mathcal P)\\
        &= - \sum_{\mathcal C^\prime \in \mathcal P^\prime } \left( \frac{g_{\mathcal C^\prime}}{v_G} \log \frac{v_{\mathcal C^\prime}}{v_G} + \sum_{x^\prime \in \mathcal C^\prime }  \frac{o_{x^\prime}}{v_G} \log \frac{o_{x^\prime}}{v_{\mathcal C^\prime}} \right) 
        \\&+ \sum_{\mathcal C \in \mathcal{P} } \left( \frac{g_{\mathcal C}}{v_G} \log \frac{v_\mathcal C}{v_G} + \sum_{x \in \mathcal C }  \frac{o_x}{v_G} \log \frac{o_x}{v_\mathcal C} \right)  \\
        &= \underbrace{ - \frac{g_{\mathcal C^\prime_j}}{v_G} \log \frac{v_{\mathcal C^\prime_j}}{v_G} - \sum_{m \in \mathcal{C}^\prime_j} \frac{o_{m}}{v_G} \log \frac{o_{m}}{v_{\mathcal C_j^\prime}}  }_{\mathcal{H}^{2}(\mathcal{C}_j^\prime)} 
           \underbrace{ -\frac{o_x}{v_G} \log \frac{o_x}{v_G}  }_{\mathcal{H}^{2}(\{x\})}  \\
        & \underbrace{ + \frac{g_{\mathcal C_j}}{v_G} \log \frac{v_{\mathcal C_j}}{v_G} + \sum_{m \in \mathcal{C}_j\setminus \{x\}} \frac{o_{m}}{v_G} \log \frac{o_{m}}{v_{\mathcal C_j}} }_{-\mathcal{H}^2(\mathcal{C}_j \setminus \{x\} )} 
               \underbrace{ + \frac{o_x}{v_G} \log \frac{o_x}{v_{\mathcal C_j}}  }_{-\mathcal{H}^2(x)} \\
        &=- \frac{g_{\mathcal C_j^\prime}}{v_G} \log \frac{v_{\mathcal C_j^\prime}}{v_G}  + \frac{g_{\mathcal C_j}}{v_G} \log \frac{v_{\mathcal C_j}}{v_G} - \frac{o_x}{v_G}(\log \frac{o_x}{v_G} - \log \frac{o_x}{v_{\mathcal C_j}} ) \\
        &- \sum_{m \in \mathcal{C}_k^\prime} \frac{o_m}{v_G} ( \log \frac{o_m}{v_{\mathcal C_j^\prime}} - \log \frac{o_m}{v_{\mathcal C_j}}) \\
        &=- \frac{g_{\mathcal C_j^\prime}}{v_G} \log \frac{v_{\mathcal C_j^\prime}}{v_G}  + \frac{g_{\mathcal C_j}}{v_G} \log \frac{v_{\mathcal C_j}}{v_G} - \frac{o_x}{v_G} \log \frac{v_{\mathcal C_j}}{v_G} \\
        &- \log \frac{v_{\mathcal C_j}}{v_{\mathcal C_j^\prime}} \sum_{m \in \mathcal{C}_j^\prime} \frac{o_m}{v_G}  \\
        &=- \frac{g_{\mathcal C_j^\prime}}{v_G} \log \frac{v_{\mathcal C_j^\prime}}{v_G}  + \frac{g_{\mathcal C_j}}{v_G} \log \frac{v_{\mathcal C_j}}{v_G} 
        - \frac{o_x}{v_G} \log \frac{v_{\mathcal C_j}}{v_G}
        - \frac{v_{\mathcal C_j^\prime}}{v_G}  \log \frac{v_{\mathcal C_j}}{v_{\mathcal C_j^\prime}}.\\
     \end{aligned}
\end{equation}

\section{Additional Details of Night-TIR Benchmark}
\label{sec:Bench}

\begin{figure}[t]
    \centering
    \includegraphics[width=1\linewidth, trim={0.2cm 0.2cm 0.2cm 0cm}, clip]{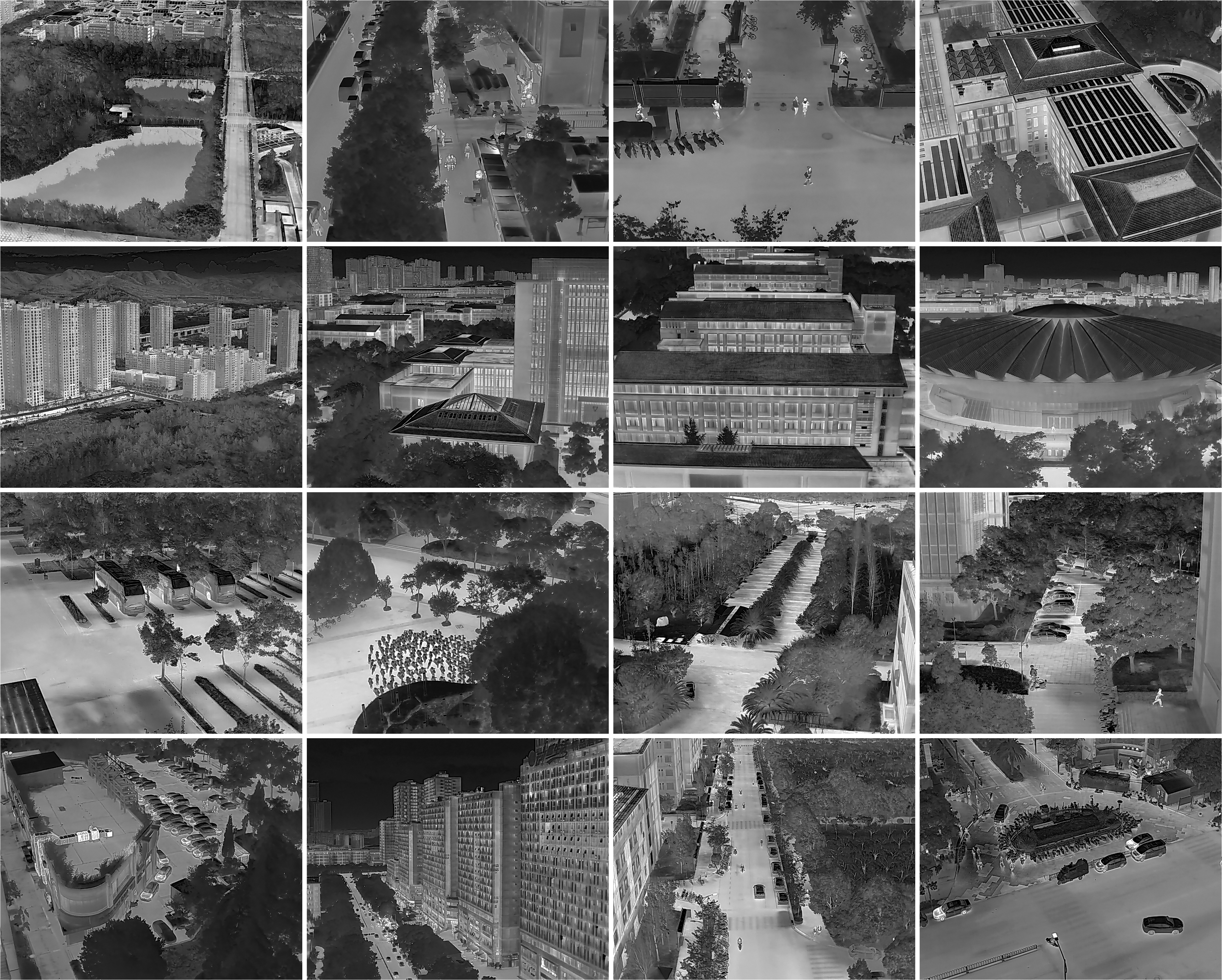}
    \vspace{-4mm}
    \caption{TIR images within Night-TIR benchmark.}
    \label{fig:night}\vspace{-3mm}
\end{figure}

Considering that nighttime scenes exhibit smaller target–background temperature differences and weaker radiative signals, thermal infrared (TIR) imagery therefore tends to have reduced contrast. 
To this end, we introduce a challenging nighttime TIR enhancement benchmark, Night-TIR.
As shown in Figure~\ref{fig:night}, Night-TIR covers a broad spectrum of nocturnal scenes—including pedestrians, vehicles, roads, woodlands, residential buildings, high-rises, plazas, lakes, and mountainous areas.
All images are captured at night using the integrated thermal camera on a DJI Matrice 4T UAV (spatial resolution $640{\times}512$, spectral band $8$–$14~\mu$m and super-resolution mode supported). 
Flights are conducted at altitudes of $20$–$80$~m; imaging is performed during hover and stabilized by a three-axis mechanical gimbal. 
To ensure data reliability, we perform rigorous manual quality control and remove low-quality samples.
In total, Night-TIR comprises $647$ high-quality nighttime TIR images, providing a realistic, diverse, and challenging benchmark for infrared image enhancement and compound-degradation restoration.

\section{Network Architectural Details }
\label{sec:Add_arch}
The proposed SEGD framework comprises an encoder, a decoder, DENet, and a set of DRMs.
The encoder contains two convolutional layers followed by eight ResBlocks (each residual block comprises two convolutional layers, each followed by GroupNorm and GELU).
The feature width is fixed at 64 channels.
The decoder consists of four ResBlocks and two convolutional layers, producing a single-channel restored image at the output.
DENet is composed of a feature extractor, a Degradation-Aware Head (DAH), and a Degradation Evidence Head (DEH).
The feature extractor mirrors the encoder architecture and appends a global average pooling layer, yielding a vector of size $(B,64)$ (where $B$ denotes the batch size). Both DAH and DEH are three-layer, fully connected multilayer perceptrons with output dimensions $(B,3)$ and $(B,2)$, respectively.
Each DRM adopts a RestormerBlock~\cite{zamir2022restormer} with 32 channels.
For DENet training, we adopt the same optimization configuration as the main backbone: the Adam optimizer with a batch size of 48; random cropping and flipping to produce $256{\times}256$ patches; an initial learning rate of $5{\times}10^{-5}$ with a cosine-annealing schedule; and 100 training epochs. After training, DENet’s parameters are frozen when training the backbone network.

\begin{figure*}[t!]
    \vspace{-2mm}
    \centering
    \includegraphics[width=0.95\linewidth, trim={0.2cm 0.2cm 0.2cm 0cm}, clip]{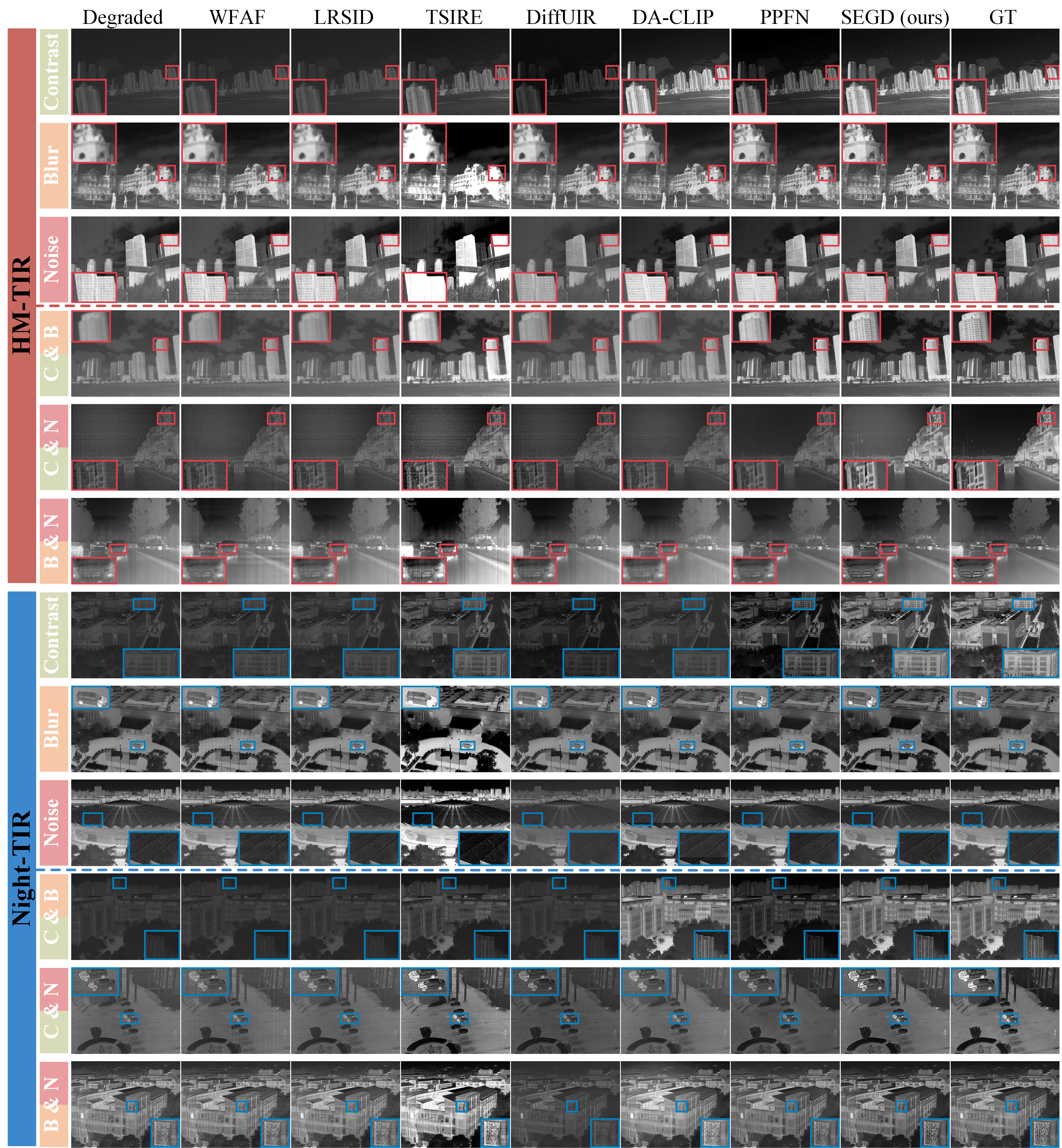}
    \vspace{-2mm}
    \caption{Additional qualitative comparison of single and double compound degradations on the HM-TIR and Night-TIR datasets.}
    \label{fig:S_D}\vspace{-4mm}
\end{figure*}
\section{More Results on HM-TIR and Night-TIR}
To thoroughly evaluate the proposed SEGD, this section presents three additional studies: (i) qualitative comparisons with competing methods under single- and double-degradation settings; (ii) comparisons between \textit{SEGD} and state-of-the-art single-degradation methods in the single-degradation regime; and (iii) evaluations against additional all-in-one visible image enhancement methods.

\label{subsec:QRSD}
\textbf{Additional qualitative comparisons.}
Figure~\ref{fig:S_D} presents qualitative comparisons between the proposed SEGD and competing methods under single- and double-degradation settings.
PPFN performs comparably to SEGD on deblurring and denoising but lags in contrast enhancement. 
Under double-degradation scenarios, SEGD yields sharper edges and more faithful repetitive textures across both datasets, effectively suppresses background noise, recovers high-frequency details, and delivers more natural contrast, resulting in clearly superior overall visual quality relative to the other methods.

\begin{figure*}[t!]
    \vspace{-2mm}
    \centering
    \includegraphics[width=0.95\linewidth, trim={0.2cm 0.2cm 0.2cm 0cm}, clip]{Pics/Single_Comp.png}
    \vspace{-2mm}
    \caption{Qualitative and quantitative comparisons under single-degradation settings on HM-TIR and Night-TIR.}
    \label{fig:Single_comp}\vspace{-4mm}
\end{figure*}

\textbf{Results on single degradation TIR}.
To assess the performance of the proposed SEGD under single-degradation settings, we construct three category-specific test subsets—noise, blur, and low contrast—on both HM-TIR and Night-TIR datasets. 
For contrast enhancement, we compare against LLFormer~\cite{wang2023ultra}, ReDDiT~\cite{lan2025efficient}, SCI++~\cite{ma2025learning}, and DarkIR~\cite{feijoo2025darkir}; for deblurring, we compare with MIMO-UNet~\cite{cho2021rethinking}, Stripformer~\cite{tsai2022stripformer}, FFTformer~\cite{kong2023efficient}, and EVSSIM~\cite{kong2025efficient}; and for denoising, we consider CycleISP~\cite{zamir2020cycleisp}, AP-BSN~\cite{lee2022ap}, DCD-Net~\cite{zou2023iterative}, and ScoreDVI~\cite{cheng2023score}.
As illustrated in Fig.~\ref{fig:Single_comp}, these methods alleviate the degradations to some extent but still suffer from issues such as insufficient contrast or local overexposure, blurred fine details, and residual noise.
In contrast, SEGD consistently delivers superior results across all three tasks—achieving the highest PSNR and SSIM—while producing sharper edges, more faithful textures, and a more natural overall appearance.

\textbf{Additional evaluation of all-in-One visible-image enhancement methods.}
As shown in Table~\ref{tab:AIO}, we additionally evaluate two recent all-in-one visible enhancement methods—DCPT~\cite{hu2025universal} and MoCE-IR~\cite{zamfir2025complexity}. 
Both methods transfer poorly to TIR images, primarily due to the pronounced modality gap in imaging physics and degradation distributions between the visible and TIR domains. These findings underscore the limited effectiveness of directly porting visible methods to TIR restoration and highlight the risk of negative transfer in such cross-domain deployments.

\begin{table*}[t]
    \centering
    \setlength{\tabcolsep}{2.7pt}
    \caption{Quantitative comparison with additional visible all-in-one methods on the  HM-TIR and Night-TIR datasets. The best and second-best performances for each metric are highlighted with \colorbox[HTML]{ffc7ce}{Red} and \colorbox[HTML]{d9e7f4}{Blue} backgrounds, respectively.}
    \small
    \label{tab:AIO}  
    \begin{threeparttable}
        \vspace{-2mm}
        \begin{tabularx}{\linewidth}{l|cc|cc|cc|cc|cc|cc|cc|cc}
        \toprule
        Dataset&\multicolumn{8}{c|}{HM-TIR}&\multicolumn{8}{c}{Night-TIR}\\
        \midrule
         Degradation&\multicolumn{2}{c|}{Single}&\multicolumn{2}{c|}{Double}& \multicolumn{2}{c|}{Triple}&\multicolumn{2}{c|}{Avg.}&
        \multicolumn{2}{c|}{Single}&\multicolumn{2}{c|}{Double}& \multicolumn{2}{c|}{Triple}&\multicolumn{2}{c}{Avg.}\\
        \midrule
         Metric&PSNR&SSIM&PSNR&SSIM&PSNR&SSIM&PSNR&SSIM&PSNR&SSIM&PSNR&SSIM&PSNR&SSIM&PSNR&SSIM\\
        \midrule
        DCPT~\cite{hu2025universal} &\footnotesize 17.878& \footnotesize \strut 0.627&\footnotesize 16.672& \footnotesize 0.529&\footnotesize \cellcolor[HTML]{d9e7f4}{\strut 16.714}&\footnotesize \cellcolor[HTML]{d9e7f4}{\strut0.326}&\footnotesize 17.088&\footnotesize 0.494&\footnotesize 19.236 &\footnotesize 0.707 &\footnotesize 16.474 &\footnotesize 0.499&\footnotesize \cellcolor[HTML]{d9e7f4}{\strut16.315} &\footnotesize \cellcolor[HTML]{d9e7f4}{\strut0.323} &\footnotesize 17.341 &\footnotesize 0.510\\
        MoCE-IR~\cite{zamfir2025complexity}&\footnotesize  \cellcolor[HTML]{d9e7f4}{\strut23.257}& \cellcolor[HTML]{d9e7f4}{\strut0.737}&\footnotesize \cellcolor[HTML]{d9e7f4}{\strut 18.061}& \footnotesize \cellcolor[HTML]{d9e7f4}{\strut 0.537}&\footnotesize 16.561&\footnotesize 0.324&\footnotesize \cellcolor[HTML]{d9e7f4}{\strut 19.292}&\footnotesize \cellcolor[HTML]{d9e7f4}{\strut 0.533}&\footnotesize \cellcolor[HTML]{d9e7f4}{22.013} & \footnotesize \cellcolor[HTML]{d9e7f4}{\strut0.746}&\footnotesize \cellcolor[HTML]{d9e7f4}{17.573}& \footnotesize \cellcolor[HTML]{d9e7f4}{0.503}&\footnotesize 16.244&\footnotesize 0.320 &\footnotesize \cellcolor[HTML]{d9e7f4}{\strut18.610} &\cellcolor[HTML]{d9e7f4}{\strut 0.522}\\
        SEGD (ours)&\footnotesize \cellcolor[HTML]{ffc7ce}{\strut 31.601}& \footnotesize  \cellcolor[HTML]{ffc7ce}{\strut 0.951}&\footnotesize  \cellcolor[HTML]{ffc7ce}{\strut 26.342}& \footnotesize  \cellcolor[HTML]{ffc7ce}{\strut 0.872}&\footnotesize  \cellcolor[HTML]{ffc7ce}{\strut 20.310}&\footnotesize  \cellcolor[HTML]{ffc7ce}{\strut 0.761}&\footnotesize  \cellcolor[HTML]{ffc7ce}{\strut 26.085}&\footnotesize  \cellcolor[HTML]{ffc7ce}{\strut 0.862}&\footnotesize  \cellcolor[HTML]{ffc7ce}{\strut 26.789} &\footnotesize  \cellcolor[HTML]{ffc7ce}{\strut 0.922} &\footnotesize  \cellcolor[HTML]{ffc7ce}{\strut 22.323} &\footnotesize  \cellcolor[HTML]{ffc7ce}{\strut 0.842}&\footnotesize  \cellcolor[HTML]{ffc7ce}{\strut 18.312} &\footnotesize  \cellcolor[HTML]{ffc7ce}{\strut 0.737} &\footnotesize  \cellcolor[HTML]{ffc7ce}{\strut 22.475} &\footnotesize  \cellcolor[HTML]{ffc7ce}{\strut 0.834}\\
        \bottomrule
        \end{tabularx}
    \end{threeparttable}\vspace{-2mm}
\end{table*}  
\section{Complexity Comparison}

For a fair comparison, we compute and report the number of learnable parameters, inference time, and floating-point operations (FLOPs) for SEGD and all competing methods on single-degradation HM-TIR inputs at a resolution of $640{\times}512$, excluding the non–deep-learning baselines WFAF and LRSID. 
As summarized in Table~\ref{tab:ComplexityC}, SEGD has the fewest parameters among all methods and achieves the second-fastest inference. 
Owing to backbone convolutions on full-resolution feature maps, SEGD does not achieve the lowest FLOPs; nevertheless, it delivers markedly superior restoration quality under complex degradations. 
Overall, SEGD offers a superior trade-off among inference efficiency, model size, and restoration quality.

\begin{table}[t]
    \centering
    \setlength{\tabcolsep}{1.6pt} 
    \caption{Complexity comparison on parameters, FLOPs, and time. The best and second-best results for each metric are highlighted with \colorbox[HTML]{ffc7ce}{Red} and \colorbox[HTML]{d9e7f4}{Blue} backgrounds, respectively.}
    \label{tab:ComplexityC}
    \footnotesize
 \begin{threeparttable}
        \vspace{-2mm}
        \begin{tabularx}{\linewidth}{l*{7}{>{\centering\arraybackslash}c}}
            \toprule
             Methods& TSIRE & DiffUIR & DACLIP & DCPT & MoCE-IR & PPFN & SEGD\\
            \midrule 
             Params (M)& \cellcolor[HTML]{d9e7f4}{\strut 2.52} &12.41 &233.14 &26.10& 23.49& 26.60 &\cellcolor[HTML]{ffc7ce}{\strut 2.268} \\
             FLOPs (G)& \cellcolor[HTML]{ffc7ce}{\strut 77.91} &494.04 &660.18 &704.95& \cellcolor[HTML]{d9e7f4}{\strut 446.41}& 704.33 & 589.48 \\
             Time (S)& \cellcolor[HTML]{ffc7ce}{\strut 0.01} &0.43 &15.32 & 0.27 & 0.32 & 0.65 & \cellcolor[HTML]{d9e7f4}{\strut 0.19}\\
            \bottomrule
        \end{tabularx}
    \end{threeparttable}\vspace{-4mm}
\end{table}

\section{Additional Ablation Studies}
\label{sec:Add_Ab}

All experiments in this section are conducted on the HM-TIR dataset.
We first examine training strategies for DENet: the DAH and DEH are trained either independently—each with its own feature extractor—or jointly with a shared extractor.
As shown in Table~\ref{tab:JT}, independent training yields a slight increase in degradation-type classification accuracy and a small reduction in the mean absolute error (MAE) of degradation-intensity estimation; however, the overall restoration quality remains essentially unchanged (PSNR increases by only 0.046~dB and SSIM is nearly identical).
Meanwhile, independent training introduces additional parameters and longer training time.

\begin{table}[b]
    \centering
    \setlength{\tabcolsep}{5.3pt} 
    \vspace{-2mm}
    \caption{Ablation of DENet Training Strategies. The best results for each metric are highlighted with \colorbox[HTML]{ffc7ce}{Red} backgrounds.}
    \label{tab:JT}
    \small
 \begin{threeparttable}
        \vspace{-2mm}
        \begin{tabularx}{\linewidth}{l|cc|c|cc}
            \toprule
             Module& \multicolumn{2}{c|}{DAH}& DEH & \multicolumn{2}{c}{SEDA}\\
            \midrule 
             Metric& EM-ACC& F1 & MAE $\downarrow$ & PSNR & SSIM\\
             \midrule 
             Independent & \cellcolor[HTML]{ffc7ce}{\strut 0.966} & \cellcolor[HTML]{ffc7ce}{\strut 0.995} & \cellcolor[HTML]{ffc7ce}{\strut0.031} &\cellcolor[HTML]{ffc7ce}{\strut 26.131} & \cellcolor[HTML]{ffc7ce}{\strut 0.862}\\
             Joint & 0.957 & 0.991 & 0.034 & 26.085 & \cellcolor[HTML]{ffc7ce}{\strut 0.862}\\
            \bottomrule
        \end{tabularx}
    \end{threeparttable}
\end{table}

We further assess how each  DRM behaves under mismatched degradations.
Concretely, we deliberately permute the mapping between degradation types and DRMs. For example, assigning the contrast DRM to handle blur or noise, and analogously for the other DRMs.
We evaluate three configurations: fully mismatched (all degradations processed by incorrect DRMs), partially matched (at least one correct DRM–degradation pair), and fully matched (one-to-one correspondence).
As summarized in Table~\ref{tab:DRM_mismatch}, the fully mismatched setting performs the worst; introducing any correct pairing produces a performance rebound; and the fully matched case achieves the best results.
These observations indicate that each DRM’s benefit is primarily specific to its designated degradation, and the gains diminish—or even vanish—when transferred across degradation types.
Notably, in triple-degradation scenarios, permuting the correspondence does not affect performance because all DRMs are invoked jointly to handle the compound degradations.
By contrast, in single- or double-degradation settings, mismatches between DRM and the degradation lead to clear performance drops.
\begin{table}[h]
    \centering
    \setlength{\tabcolsep}{4pt} 
    \caption{Impact of DRM-degradation assignment on performance. The best and second-best results for each metric are highlighted with \colorbox[HTML]{ffc7ce}{Red} and \colorbox[HTML]{d9e7f4}{Blue} backgrounds, respectively.}
    \label{tab:DRM_mismatch}
    \small
 \begin{threeparttable}
        \vspace{-2mm}
        \begin{tabularx}{\linewidth}{l*{5}{>{\centering\arraybackslash}c}}
            \toprule
             & \multirow{2}{*}{\makecell{Contrast\\Enhancement}} & \multirow{2}{*}{Deblurring} & \multirow{2}{*}{Denoising} & \multicolumn{2}{c}{Avg. Metrics} \\
            \cmidrule{5-6}
             & & & & PSNR & SSIM\\
            \midrule 
             1 & NDRM & CDRM& BDRM & 21.064&0.801 \\
             2 & CDRM & NDRM & BDRM & \cellcolor[HTML]{d9e7f4}{\strut 24.771} & 0.\cellcolor[HTML]{d9e7f4}{\strut 836}\\
             3 & NDRM & BDRM & CDRM & 21.816 & 0.807\\
             4 & BDRM & CDRM & NDRM & 23.275 & 0.833\\
             5 & CDRM & BDRM & NDRM & \cellcolor[HTML]{ffc7ce}{\strut 26.085} & \cellcolor[HTML]{ffc7ce}{\strut 0.862}\\
            \bottomrule
        \end{tabularx}
    \end{threeparttable}\vspace{-2mm}
\end{table}

Finally, we validate the effectiveness of the 2D-SE–based weighted aggregation in SE-ROS by replacing it with simple averaging for the selected features.
Because averaging ignores differences in feature importance, performance in the triple-degradation setting declines slightly, with PSNR/SSIM reduced by 0.212~dB and 0.007, respectively.

\section{Limitations and Future Work}
We follow the degradation synthesis strategy designed in PPFN~\cite{liu2025enhancing}, where TIR degradations are categorized into low contrast, blur, and noise, and training samples are generated accordingly. However, as noted in PPFN, obtaining strictly paired degraded--clean TIR images is inherently difficult, and any degradation pipeline constructed from limited priors can hardly cover the full range of more complex and diverse real-world degradations. Consequently, when encountering degradation types or atypical combinations that lie outside the training distribution, the performance of the model may still deteriorate to some extent. Nevertheless, experiments on real-world datasets such as AWMM demonstrate that the proposed SEGD still exhibits satisfactory generalization under no-reference quality metrics.

From a design standpoint, SEGD is highly extensible: to support additional degradation types, one can simply append the corresponding DRMs without modifying the overall framework.  However, increasing the number of degradation types also introduces a non-negligible cost: as the number of types grows, the candidate restoration paths expand combinatorially, which in turn increases the number of trainable parameters and the training time. By contrast, the impact on the SE-ROS strategy is relatively minor, since the underlying linear-time 2D-SE minimization algorithm~\cite{xian2025community} incurs very low inference overhead on small graphs, and all quantities admit dynamic updates. Thus, we can perform constant-time updates of each node’s 2D-SE contribution.

Overall, compared with visible-image enhancement, TIR enhancement is equally important yet remains relatively underexplored.  In future work, we plan to: (i) construct more comprehensive and physically consistent TIR degradation models so that synthesized training data better cover real-world degradations; and (ii) develop restoration paradigms that accommodate unknown or even open-world degradations, enabling stable and reliable performance under extreme or previously unseen conditions.

\end{document}